# Themes of Revenge:

# Automatic Identification of Vengeful Content in Textual Data

Yair Neuman[1]* Eden Shalom Erez[2] Joshua Tschantret[3] Hayden Weiss[4]

[1] The Department of Cognitive and Brain Sciences and the Zlotowski Center for Neuroscience, Ben-Gurion University of the Negev, Beer-Sheva, Israel

yneuman@bgu.ac.il

[2] Independent Researcher, edenerez@gmail.com

[3] Department of Political Science, Emory University, 1555 Dickey Drive, Atlanta, GA 30322. E-mail: joshua.tschantret@emory.edu

[4] Byram Hills High School. 12 Tripp Ln, Armonk, NY 10504

weissh22@byramhills.net

*Corresponding author

**Abstract**

Revenge is a powerful motivating force reported to underlie the behavior of various solo perpetrators, from school shooters to right wing terrorists. In this paper, we develop an automated methodology for identifying vengeful themes in textual data. Testing the model on four datasets (vengeful texts from social media, school shooters, Right Wing terrorist and Islamic terrorists), we present promising results, even when the methodology is tested on extremely imbalanced datasets. The paper not only presents a simple and powerful methodology that may be used for the screening of solo perpetrators but also validate the simple theoretical model of revenge.

# Themes of Revenge:

# Automatic Identification of Vengeful Content in Textual Data

**Introduction**

Revenge is a prominent theme in the narrative arts across time and space. It is explored in works from Homer to Shakespeare, a ubiquity that indicates the power of revenge as a motivating force in human conduct. The films *Inglorious Basterds*[1] and *Django unchained*[2] are just two recent examples illustrating the ubiquity of the revenge theme in cinema, literature, and theatre. While the theme of revenge is intensively represented in the arts, it is argued that revenge receives relatively limited attention in the psychological research (Jackson, Choi & Gelfand, 2019). This minimal attention is surprising as recent research in areas of great public interest—from terrorism (Tschantret, 2020) and international aggression (Stein 2015) to revenge pornography (Dymock & Van Der Westhuizen, 2019) and school shooting (Neuman, Lev Ran & Erez, 2020)—highlight the relevance of revenge and invite new directions for studying it, specifically in practical contexts of screening for revenge motivated perpetrators.

While the psychology underlying revenge has been studied and debated for years (e.g. Kohut, 1972; Neuman, 2012), there seems to be an emerging and converging understanding of the phenomenon (Jackson, Choi & Gelfand, 2019), at least with regards to its constitutive dimensions. First, it must be acknowledged that revenge is a multidimensional phenomenon rather than the simple outcome of a narcissistic injury, as believed in the past (Kohut, 1972). As such, revenge may emerge from several sources but, from a psychological perspective, it seems to be a response to a harmful action which is conceived as morally reprehensible or norm violating (e.g., Bavik &

---

[1] https://www.rottentomatoes.com/m/inglourious_basterds
[2] https://www.imdb.com/title/tt1853728/

Bavik 2015, Carlsmith et al. 2002). Such moral wrongdoing, which is experienced as humiliating, may be conceptualized in terms of narcissistic rage resulting from narcissistic injury (Akhtar, 2018). This rage is accompanied by fantasies of aggressive retaliation that can be described as sadistic (Buckels, Jones& Paulhus, 2013; Foulkes, 2019; Plouffe, Smith & Saklofske, 2019) because they involve the imagined pleasure of getting even and restoring a lost equilibrium between the avenger and his object of retaliation (Neuman, 2012). Given this scenario, we may introduce a prototypical narrative of revenge. It starts with a first-person experience of humiliation. The humiliation is framed within a larger context of injustice representing threat to order and danger to the self. As the subject feels defenseless and lonely, a foundational aspect of his self is threatened (Holmes, 2011) in a way that generates fantasies of getting even by gaining pleasure through the humiliation of the aggressor. This is the model and the narrative that we adopt in our study for operationalizing and automatically measuring dimensions of revenge.

An open question, which is the primary focus of our study, is whether advances in Natural Language Processing and Machine Learning can be used for automatically identifying themes of revenge in textual data. We use the term "themes" in the sense of a topic of discourse, as we are specifically interested in a topical analysis of a text. It goes without saying that the ability to automatically identify themes of revenge may have clear practical benefits: From the screening of white supremacists expressing vengeful intentions that may be used as warning signals for violent actions, to the identification of disgruntled employees who may pose the "insider threat". *To the best of our knowledge, the automatic identification of "vengeful texts" has never been the focus of a research paper*. The aim of the current paper is to test a simple and straightforward approach for the automatic identification of vengeful themes in textual data. When tested on several datasets, this methodology shows preliminary but promising results.

Our methodology shows preliminary but promising results when tested on several datasets. We demonstrate an ability to classify vengeful texts out-of-sample in both balanced and unbalanced datasets, the latter being more commonly encountered in real-world scenarios. An additional analysis shows that the methodology can be used to identify terrorist-authored writings well above baseline probabilities based on vengeful content. Lastly, an illustrative analysis compares two screenplays based on the dialogue of two archetypal vengeful characters. Taken together, the results reveal the strength of theoretically motivated machine learning in classifying vengeful texts and identifying vengeful actors.

**2. Materials and methods**

**2.1. Data**

For the first analysis, personal narrative texts (N =100) were manually collected from several sources, most prevalently revenge-based subreddits and text identified in 4chan, 8kun, and independent blogs. Texts were selected after careful analysis assuring that a vengeful theme is present within each text. For a comparative analysis, and following previous works (Neuman et al., 2015; Tschantret, 2020), we also used the texts from the Blog Authorship Corpus (Schler et al., 2006). For maximizing the overlap with previous studies using this corpus in the context of solo perpetrators, we used the texts of male bloggers only, aged between 15 and 25. The texts written by each subject were merged into a single file. Overall, the final dataset of the bloggers that we have used for the analysis featured the texts written by 5029 subjects.

**2.2. Pre-processing**

Texts written by the same author, have been collapsed into a single document. Using SpaCy[3], each document has been segmented to sentences and POS tagging was used to identify only nouns, verbs, adjectives and adverbs. A whole word was counted only if it appeared in the Word2Vec dictionary of Gensim[4]. Otherwise, we used the word's lemma if it appeared in the dictionary. Texts including less than 100 words were removed from the analysis leaving us with N = 83 for the group of "vengeful documents" and N = 4829 of non-vengeful documents gained from the Bloggers Corpus.

Next, we computed the tf-idf score for each word in a document, ranked the words and chose the top-ranked 100 words for each document separately. This method is best representing the document. In this way, we have intermediately represented each document as a vector of length 100 regardless of the document's original length.

**2.3. Measuring themes of revenge**

For measuring the extent in which themes of revenge appear in a document, we have used the vectorial semantics approach to personality analysis (Giachanou et al., 2020; Neuman, 2016; Neuman & Cohen, 2014; Tschantret, 2020). To repeat and based on integrative analysis of the psychological research dealing with revenge, we considered revenge to be a situation where an individual experiences *Narcissistic Injury* as expressed by a deep sense of *Humiliation*. The situation is conceived as one of *Injustice* and as violating *Moral* norms. As the individual experiences *Loneliness* and cannot digest his mental pain through the support of others, he responds with *Narcissistic Rage*, expressed in *sadistic* fantasies where *pleasure* is experienced from *humiliating* the offender and getting even. Based on this theoretical based conceptualization of revenge, we have defined the following dimensions of revenge: (1) Humiliation as experienced firsthand from the avenger's perspective (2) Loneliness (3), Unjust, and (4) Sadism (Humiliation of the offender + Pleasure).

---

[3] https://spacy.io/
[4] https://radimrehurek.com/gensim/

These dimensions are operationalized through vectors of words that aim to represent their meaning. For building the vectors, we have used *The iWeb Corpus*[5] that contains 14 billion words from 22 million web pages. We have used eight seed words corresponding with the above dimensions: Unjust, Unfair, Revenge, Immoral, Lonely, Pleasure, Humiliate, and Humiliated. Searching the words collocated with the target seed in a lexical window of plus/minus 4 words and setting a selection criterion of Mutual Information >= 3, we have identified the words collocated with each target seed. Only words whose frequency in the corpus was => 10 were selected. The output of such a search is a list of words classified and presented according to the POS tags: +Adj, +Verb, +Noun and +Adv. We selected words for the vectors separately from each of these categories and decided which words to include based on the expert's judgment of the first author. For example, the dimension of Revenge has been operationalized in three different vectors of words. One for adjectives, one for verbs and one for nouns:

Revenge_Adj: Sweet, bloody, violent …
Revenge_Verb: Seek, kill, plot …
Revnge_Noun: Attack, desire, kill …

---

[5] https://www.english-corpora.org/iweb/

Through expert's analysis, adverbs associated with "revenge" were found to be of a minor value. Each of the above vectors contained 99 words, but the different vectors that we have used varied in their length. The final vectors were sets of collocations under the titles of: (1) Revenge (Adj) (2) Revenge (Verb), (3) Revenge (Noun), (4) Immoral (Adj), (5) Lonely (Adj), (6) Pleasure (Adj), (7) Humiliate (Verb), (8) Unfair (Adj), (9) Humiliated (Verb) and (10) Humiliated (Adj).

For measuring the degree in which a document expresses themes of revenge, we have used the algorithm of Word2Vec (Mikolov et al., 2013), trained on Google News, and measured the degree of association between each of the 100 words representing the document and each of the words in a defined vector. The final degree of association between a text and a vector has been measured in two different ways: either by averaging across words or by identifying the maximal degree of association found between a word in the document and a word in the vector, where the similarity is defined as: Sim(word_1, word_2) = 1-cos(word2vec[word_1], word2vec[word_2]).

The similarity of each document to each vector has been normalized (0-1) using the MinMax scaling method, where the similarity score of the document to the specific revenge vector has been computed as follows:

$$\frac{Score - MinScore}{MaxScore - MinScore}$$

We aim to classify documents as vengeful given their semantic similarity to the abovementioned dimensions of revenge. If this task could have been successfully achieved, then a clear side benefit of the current study is the validation of the vengeful dimensions as discussed in the literature and presented in the abovementioned model.

## 3. Analysis and Results

We present several analyses in an increasing order of complexity. Our first analysis aimed to test the ability to classify vengeful vs. non-vengeful texts by using the association/similarity scores. We first compared the similarity scores of the vengeful (N = 83) vs. non-vengeful documents (N = 4829) using (1) a Bayesian T-Test for Independent Samples and (2) Independent Samples T-Test. Both tests were run through JASP[6]. For identifying the most important features differentiating between the two classes, we have used two measures: (1) the Bayes Factor and (2) the Vovk-Sellke Maximum p-Ratio. The most distinguishing features representing each dimension, were found to be the following:

1. Revenge (Adj, Verb, Noun) for the max score
2. Unjust (Adj)
3. Immoral (Adj)
4. Lonely (Adj)
5. Humiliate (Verb) for the max score
6. Unfair (Adj)
7. Humiliated (Verb, Adj)

These dimensions clearly support the revenge model through which the dimensions of revenge have been operationalized.

### 3.1. Analysis 1.

For the first classification task, we have formed a balanced dataset of 83 documents tagged as vengeful and a random sample of 83 documents from the Blogs Corpus tagged as non-vengeful. For the first analysis we have used JASP and tried to classify the documents using their similarity scores as features.

---

[6] https://jasp-stats.org/

A Boosting Classification model with tenfold cross-validation performed well with 70% Precision, 88% Recall and 87% of AUC. The AUC curve appears in figure 1:

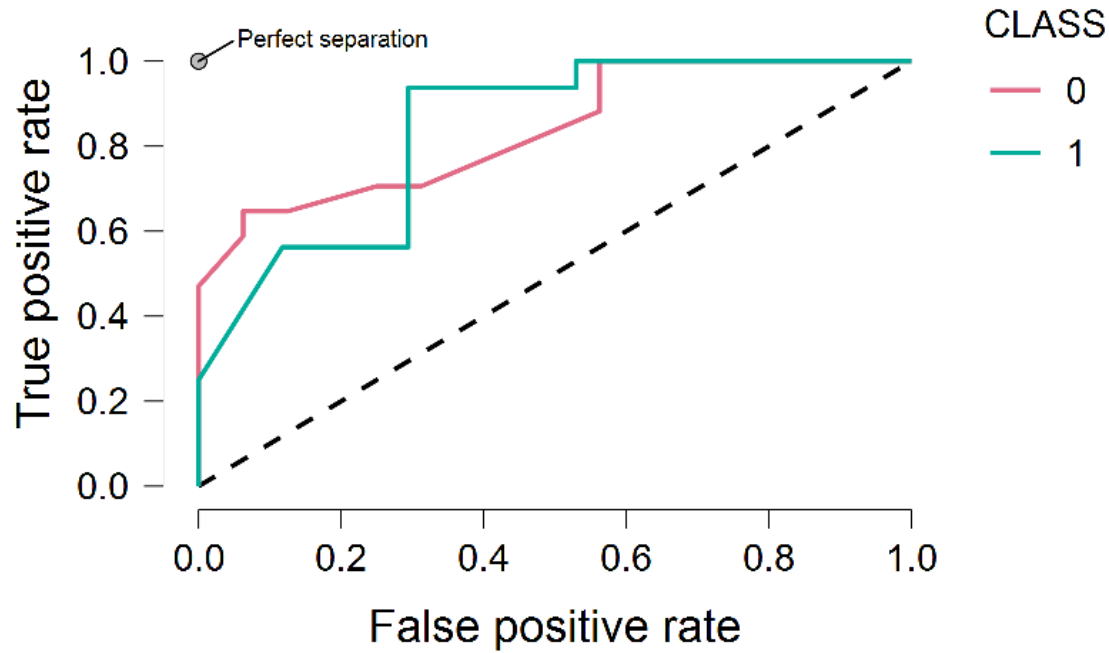

Figure 1. The AUC Curve

The three features best contributing to the classification were: Unfair, Revenge (Noun) and Revenge (Adj). For increasing the validity of the results, we have repeated the analysis with three other classifiers and variations of the sampling methods: Random Forest and Linear Discriminant, with 20% of the data for the validation and 20% of the data for test, and K-NN with 10-fold cross validation. The results are presented in the next table:

|  | Precision | Recall | AUC |
|---|---|---|---|
| **Random Forest** | 88 | 88 | 94 |
| **Linear Discriminant** | 94 | 100 | 97 |
| **K-NN** | 61 | 79 | 91 |

Table 1. Performance of the classifiers

We can see that across the different variations of the ML classifiers, the results are satisfactory.

The above results might be qualified by arguing that the existence of the word "revenge" and its forms in the text, is a simple indication of a vengeful theme, and no need exists for using of the vectors and their similarity to the text. To address this critique, we have removed the word "revenge" and its forms from the vengeful documents and repeated the analysis. This time, and using the criterion of the Bayes Factor, it was found that only one of the Revenge measures – Revenge (Adj) differentiated between the classes. We repeated the same ML procedure, this time with Revenge (Adj) as the only feature representing the Revenge vectors. The results are presented in the next table:

|  | Precision | Recall | AUC |
|---|---|---|---|
| **Boosting classification** | 57 | 67 | 70 |
| **Random Forest** | 73 | 84 | 85 |
| **Linear Discriminant** | 80 | 92 | 97 |
| **K-NN** | 76 | 76 | 76 |

Table 2. The performance of the classifiers when the word "revenge" has been removed

As expected, performance deteriorated as a result of deleting a crucial keyword and as a result of the significant reduction in the number of features. However, even in this case a "lazy" classifier such as K-NN tested with tenfold cross-validation showed a high level of performance with 76% Precision/Recall. It means that 76% of the documents with a vengeful signature have been correctly identified. We may conclude that the simple vectorial approach used in this study can produce good classification performance even when the most significant keyword is removed from the documents.

**3.2. Analysis 2**

Up to now, we have used a balanced dataset that included an equal number of vengeful and non-vengeful documents. However, it is interesting to test the performance of classifiers working on an unbalanced dataset where the vengeful documents have a very low prevalence. This situation is closer to real life situations where we would like to screen for potential perpetrators motivated by revenge. For addressing this challenge, we have formed a single integrated file of the vengeful texts and the bloggers texts, where the prevalence of the vengeful texts was extremely low: 1.7%. This is a difficult situation of finding a needle-in-a-haystack. For addressing this challenge, we have used the dataset that includes the vengeful texts ($N = 83$) and all the non-vengeful texts ($N = 4829$).

Two ML classifiers trained through a weighted training procedure have gained the best classificatory results. AdaBoost Tree gained 34% Precision and 70% Recall, and a Gradient Boosting Classifier gained 49% Precision and 67% Recall. The rounded percent of vengeful documents in our dataset is 2%. Therefore, trying to randomly "hit" a vengeful text, would be highly unlikely. What are the chances that a document is vengeful given that the ML classifier has tagged it as such? Around 40%, as illustrated by the average precision of the above classifiers. This is a significant improvement over the base rate of vengeful documents and the performance is specifically encouraging given the needle-in-the-haystack context of the analysis. Moreover, in such situations, the recall measure may be more important than the precision measure, as we seek to identify the maximal number of texts in order to reach potential perpetrators. As the above results show, we could have identified most of the vengeful texts regardless of their low prevalence in the dataset. Here we don't get into the price of false positives as discussed by Neuman, Cohen and Neuman (2019), but stick to the minimal level of performance expressed by the measures of precision and recall.

**3.3. Analysis 3**

We have further tested our ability to identify vengeful documents by using three additional datasets. The first dataset (Neuman, Lev-Ran & Erez, 2020) among the three includes texts written by school shooters (N =17). The second and third datasets have been produced by right wing American terrorists (N = 11) and Islamic terrorists (N = 11) and adapted from Tschantret (2020). For each analysis, we have applied the same procedure described above and used the weighted training classifier on a merged file that includes the bloggers and one of the three groups (shooters, right wing terrorists and Islamic terrorists). The next table presents the performance of the best classifier as examined through the measure of f1 macro:

| Group | ML CLASSIFIER | Recall | Precision | tp | fn | fp | tn | Base-rate |
|---|---|---|---|---|---|---|---|---|
| RW | Random Forest | 36% | 5% | 4 | 7 | 79 | 4750 | 0.2% |
| ISLA | AdaBoost | 18% | 40% | 2 | 9 | 3 | 4826 | 0.2% |
| SHOT | AdaBoost | 60% | 49% | 56 | 27 | 59 | 4770 | 0.3% |

Table 3. The performance of the classifiers when applied to the three datasets

* RW: right wing terrorists, ISLA: Islamic terrorists, SHOT: school shooters, tp: true positive, fn: false negative, fp: false positive, tn: true negative, Base-rate: the prevalence of the group in the dataset.

We can see that the best classificatory results have been gained for the texts written by school shooters, where the classifier identified 60% of the texts. This result may point to the importance of vengeful intentions among school shooters, a finding that corresponds with both qualitative and quantitative analysis of shooters (Knoll, 2010; Neuman et al., 2015). The precision of the classifier was also high: in 49% of the cases where the classifier tagged the text as vengeful it was a text produced by a shooter.

While the results are not as strong in the case of terrorist-authored documents, the optimal ML algorithms still show an impressive ability to classify both right-wing and Islamist texts based strictly on vengeful themes. 36% and 18% of the right-wing and Islamist terrorist texts were accurately identified, respectively. Similarly, as the precision values indicate, the chances that a text is written by a right-wing or Islamist given that the classifier identifies it as vengeful are 5% and 40%, respectively. Given the very low base rates (0.2%), which mirror the low proportion of terrorists even among subpopulations sympathetic to radical views, these results show a promising ability to classify potentially violent extremists of various ideological stripes. Not only are these results consistent with previous research that reveals a preoccupation with vengefulness in terrorist writings and improved classification when incorporating these themes into ML models (Tschantret, 2020), but they also highlight the benefit of theoretically motivated analysis in ML research. Scholars have long theorized that terrorists are often driven by vengefulness (Crenshaw, 1981; Post, 2007). Lacking appropriate methodologies, however, this connection has eluded empirical testing and application to real-world scenarios.

As we have opened the paper by mentioning Django unchained, we were curious to examine the extent in which themes of revenge exists in the texts produced by Django. For a comparative analysis, we have used the script of Joker[7] where the character Arthur Fleck (the Joker) is similar to Django in its violent behavior but different in its psychological signature. In both cases, we have a violent character, but Django is conceived to be much more motivated by revenge. We analyzed the texts produced by these two characters and measured its similarity to the dimensions of revenge as discussed above. This analysis is illustrative only. The results are presented in table 4.

| Dimension | Django | The Joker |
|---|---|---|
| **Revenge Adj** | 0.56 | 0.34 |
| **Revenge Verb** | 0.57 | 0.35 |
| **Revenge Noun** | 0.41 | 0.37 |
| **Unjust** | 0.40 | 0.18 |
| **Immoral** | 0.54 | 0.28 |
| **Lonely** | 0.56 | 0.49 |
| **Humiliate** | 0.59 | 0.40 |
| **Unfair** | 0.35 | 0.37 |
| **Humiliated** | 0.52 | 0.42 |

Table 4. Association between the characters' texts and the dimensions of revenge

This table shows that the dimensions of revenge are much more salient in the character of Django than in the character of the Joker, as expected. As we have opened with a reference to *Django unchained* as a movie epitomizing the theme of revenge, the above results may serve as an illustrative closure.

---

[7] https://www.imdb.com/title/tt7286456/

## 4. Discussion and Conclusions

Prediction in the social sciences is a difficult task with surprisingly little success (e.g., Salganik et al., 2020). In this context, we must frame our results to avoid any pretentious and ungrounded arguments. The context of the current study is the identification of vengeful themes in textual data. While the results are limited to specific datasets, they are encouraging specifically as they were gained through out-of-sample validation methodologies, by using four different datasets, and as they are firmly grounded in deep theoretical analysis rather than a result of a bottom-up approach. The last point is of a specific interest as the black box dynamics of supervised Machine Learning may produce convincing classification results that cannot be easily grounded in theoretical understanding. Our approach leans on the extent to which a text is semantically similar to contexts where well defined dimensions of revenge are evident. As such, the benefits of the proposed approach, is that it is theory motivated, easy to understand and much easier to implement than a dictionary-based approach (e.g., van der Vegt et al., 2020). Moreover, our findings empirically support a simple and integrative psychological model of revenge. In this sense, the automatic methodology developed in this paper recursively and empirically support the theoretical model through which it has been designed. In sum, the paper presents a way of automatically measuring themes of revenge in textual data, seems to validate dimensions of revenge as known in the literature, support a simple and integrative model of revenge and may be applied through a careful process to the task of screening for solo and revenge motivated perpetrators.

## Appendix 1. The detailed methodology

**Data.** Each empirical section of the manuscript involves a different dataset. For analysis 1, we gathered 100 posts from revenge-theme message boards, including 4Chan and 8kun. Each post was read to ensure prima facie vengeful content that our method seeks to automatically identify from a large textual corpus. In the second analysis, these data are combined with a sample of blog posts from the Blog Authorship Corpus. Since our theoretically selected texts by school shooters and terrorists are all written by males, we restrict to blog data authored by men as well. Various texts are used in analysis 3. School shooter texts, personal documents written by the shooters themselves, were gathered from a website dedicated to the subject and that maintains primary source documents (www.schoolshooters.info/original-documents); 18 first-person accounts were obtained. Terrorist data is comprised of 11 Islamist terrorist texts and 11 right-wing terrorist texts. Right-wing terrorists—those identified as white supremacists, anti-abortionists, Christian fundamentalists, or antigovernment right-libertarians—were identified from the Global Terrorism Database and was restricted to incidents in the United States, Canada, and Western Europe. We uncovered 12 written, English-language documents, primarily manifestos, written by individuals who personally carried out terrorist attacks (e.g., Timothy McVeigh). Our collection of 12 Islamist terrorist texts is more heterogenous and includes, in addition to manifestos by individual terrorists, documents written by leaders such as Osama bin Laden's "Declaration of Jihad against the American People." In our final, illustrative analysis, we select main character dialogue from two screenplays: Joker and Django Unchained.

**Text Processing.** Texts written by the same author are combined into single texts to ensure statistical independence among observations in our dataset. We then use a part of speech tagger to identify and include in the analysis only nouns, verbs, adjectives, and adverbs. Words in each document are lemmatized to remove inflectional endings to ensure that variations of the same word are reduced to their base form. Next, the term frequency—inverse document frequency (tf-idf) is calculated for each word in a document. TF-IDF is computed as the product of term frequency, $f_{t,d}$ for term $t$ and document $d$, and inverse document frequency, $\log\left(\frac{N}{|\{d \in D: t \in d\}|}\right)$ where $N$ is the total number of documents and $|\{d \in D: t \in d\}|$ is the number of documents where term $t$ is included. A high tf-idf score therefore indicates a strong relationship with the word assigned the score and the document in which it appears. Terms within each document are ranked according to these tf-idf scores and the top 100 are selected, resulting in each document being reduced a vector of length 100 irrespective of its original length. For this reason, texts that are under 100 words in length are omitted from the analysis (eliminating one right-wing and one Islamist terrorist text).

**Measuring Revenge.** Based on our theory of revenge, we represent the constituent facets of revenge—revenge, immoral, lonely, pleasure, unfair, and humiliate—as word vectors that represent the meaning of the underlying feature. A theoretical assumption we rely upon here is the distributional hypothesis, which posits that words used in the same contexts share similar meanings (e.g., words used in the same context as 'revenge' are likelier to be conceptually similar to revenge). To build the vectors, we use The iWeb Corpus, which contains over 14 billion words draws from 22 million web pages and includes the context in which the words are used (e.g., sentences with the target word highlighted). Only words that appear at least 10 times in The iWeb Corpus are included to eliminate very rare or idiosyncratic words. We search for words collated with the target words within a ±4 lexical window. An additional selected criterion is that we restrict to words with Mutual Information ≥ 3, where Mutual Information is measures the relationship between two simultaneously sampled random variables and is calculated as $I(X;Y) = \sum_{x \in X} \sum_{y \in Y} P(x,y) \log \frac{P(x,y)}{P(x)P(y)}$. We use Word2Vec, a neural network-based word embedding algorithm, trained on Google News to measure the semantic similarity of each of the 100 words in our documents our revenge vectors. Similarity is measure with the cosine similarity metric, where $similarity(word_1, word_2) = 1 - \cos(Word2Vec[word_1], Word2Vec[word_2])$. Lastly, we normalize the similarity score using the MinMax scaling method where $\frac{Score - MinScore}{MaxScore - MinScore}$.

**Classification Analyses.** Our interest is in establishing whether revenge vectors aid in accurately classifying vengeful texts. Classification, in contrast to statistical significance testing, involves identifying classes (e.g., vengeful versus non-vengeful texts) based on whether the predicted probability of being from the substantively interesting class (e.g., vengeful text) is above a predefined classification threshold. Our sole predictive features are the scores detailed in the previous section. Several metrics are used to evaluate classification performance. We rely especially on recall, the fraction of the theoretically relevant class identified (TP/TP+FN, where TP is true positive and FN is false negative), and precision, the (TP/TP+FP) fraction of the theoretically relevant class among all those classified as such. We also evaluate the area under the receiver operating characteristic curve (AUC), which plots true positive rate (another term for recall) against the false positive rate as the classification threshold is allowed to vary from 0 to 1; the AUC is the value obtained by integrating over the curve.

Each classifier used represents a different 'classification rule' used to assign the predicted probability of observations that are compared against the classification threshold. As our goal is to optimize accurate classification of vengeful texts, we test a variety of different classifiers to determine the most successful. Each model is estimated with ten-fold cross-validation; this process involves partitioning the dataset into ten equally sized subsamples, where each tenth serves once as the test dataset on which the classifier trained on the remainder of the dataset is evaluated. ML algorithms we tested include: k-nearest neighbors, a naïve classification rule that assigns observations to the same class the majority class among the n closest observations in feature space; random forest, an ensemble of decision trees; AdaBoost, an ensemble method that aggregates results from several weak learners to produce a strong learner; linear discriminate analysis, a linear combination method whose process involves dimensionality reduction; another ensemble decision tree algorithm.

AdaBoost Tree gained 34% Precision and 70% Recall, and a Gradient Boosting Classifier gained 49% Precision and 67% Recall

**Authors' contributions**

Conceived and designed the experiments; YN, EE

Performed the experiments; YN, EE

Analyzed and interpreted the data; YN, EE, JT

Contributed reagents, materials, analysis tools or data; YN, JT, HW

Wrote the paper; YN, EE, JT, HW

**Acknowledgements**